\documentclass[journal,twoside,web]{ieeecolor}
\usepackage{tmi}
\usepackage{cite}
\usepackage{amsmath,amssymb,amsfonts}
\usepackage{graphicx}
\usepackage{textcomp}
\usepackage{subfig}
\usepackage{dsfont}
\usepackage{makecell}
\usepackage{color, soul}
\usepackage{hyperref}
\usepackage{multirow}
\usepackage{algorithmicx,algpseudocode,algorithm}

\makeatletter
\usepackage{xspace}
\def\@onedot{\ifx\@let@token.\else.\null\fi\xspace}
\DeclareRobustCommand\onedot{\futurelet\@let@token\@onedot}
\newcommand{\eqnref}[1]{Eq\onedot~\eqref{#1}}
\newcommand{\figref}[1]{Fig\onedot~\ref{#1}}
\newcommand{\algoref}[1]{Alg\onedot~\ref{#1}}

\def\BibTeX{{\rm B\kern-.05em{\sc i\kern-.025em b}\kern-.08em
    T\kern-.1667em\lower.7ex\hbox{E}\kern-.125emX}}
\begin{document}
\title{SPIRiT-Diffusion: Self-Consistency Driven Diffusion Model for Accelerated MRI}
\author{Zhuo-Xu Cui, Chentao Cao, Yue Wang, Sen Jia, Jing Cheng, Xin Liu, Hairong Zheng, \IEEEmembership{Senior Member, IEEE}, Dong Liang, \IEEEmembership{Senior Member, IEEE}, Yanjie Zhu, \IEEEmembership{Senior Member, IEEE}
\thanks{Zhuo-Xu Cui, Chentao Cao and Yue Wang contributed equally to this manuscript.}
\thanks{Corresponding author: yj.zhu@siat.ac.cn (Yanjie Zhu) and dong.liang@siat.ac.cn (Dong Liang).}
\thanks{Chentao Cao, Jing Cheng, Sen Jia, Xin Liu, Hairong Zheng, Dong Liang and Yanjie Zhu are with Lauterbur Research Center for Biomedical Imaging, Shenzhen Institute of Advanced Technology, Chinese Academy of Sciences, Shenzhen, China.}
\thanks{Zhuo-Xu Cui and Dong Liang are with the Research Center for Medical AI, Shenzhen Institutes of Advanced Technology, Chinese Academy of Sciences, Shenzhen, China.}
\thanks{Yue Wang is with the School of Biomedical Engineering, Shenzhen University Medical School, Shenzhen University,  Shenzhen, China.}
}
\maketitle

\begin{abstract}
Diffusion models have emerged as a leading methodology for image generation and have proven successful in the realm of magnetic resonance imaging (MRI) reconstruction. However, existing reconstruction methods based on diffusion models are primarily formulated in the image domain, making the reconstruction quality susceptible to inaccuracies in coil sensitivity maps (CSMs). $\textit{k}$-space interpolation methods can effectively address this issue but conventional diffusion models are not readily applicable in $\textit{k}$-space interpolation. To overcome this challenge, we introduce a novel approach called SPIRiT-Diffusion, which is a diffusion model for $\textit{k}$-space interpolation inspired by the iterative self-consistent SPIRiT method. Specifically, we utilize the iterative solver of the self-consistent term (i.e., $\textit{k}$-space physical prior) in SPIRiT to formulate a novel stochastic differential equation (SDE) governing the diffusion process. Subsequently, $\textit{k}$-space data can be interpolated by executing the diffusion process. This innovative approach highlights the optimization model’s role in designing the SDE in diffusion models, enabling the diffusion process to align closely with the physics inherent in the optimization model-a concept referred to as model-driven diffusion. We evaluated the proposed SPIRiT-Diffusion method using a 3D joint intracranial and carotid vessel wall imaging dataset. The results convincingly demonstrate its superiority over image-domain reconstruction methods, achieving high reconstruction quality even at a substantial acceleration rate of 10.

\end{abstract}


\section{Introduction}
\IEEEPARstart{M}{agnetic} resonance imaging (MRI) is widely used in clinical and research domains. However, its long acquisition time is still a major limitation, resulting in trade-offs between spatial and temporal resolution or coverage. As a result, there is a growing interest in reconstructing high-quality MR images from a limited amount of $k$-space data to accelerate acquisitions. The advances of compressed sensing (CS)\cite{lustig2008compressed, majumdar2015improving, ye2019compressed, haldar2010compressed, zhao2012image, Low-Rank} and deep learning (DL)\cite{yang2018admm, modl, zhu2018image, liang2020deep, peng2022deepsense} have made substantial progress in MR reconstruction field over the past few decades. They leverage image prior information to create hand-crafted or learnable regularizations, which are combined with a data consistency term to solve the ill-posed inverse problem of MR reconstruction.

Recently, score-based diffusion models\cite{score-based, score-based-SDE, song2020improved, DDPM} have emerged as a powerful deep generative prior in MR reconstruction\cite{song2022solving, chung2022score, jalal2021robust}. These models rely on forward and reverse stochastic differential equations (SDEs) to encode and decode (generate) images. The forward SDE includes a drift term representing the deterministic trend of the forward process and a diffusion term introducing random fluctuations. For instance, the drift term in Variance Preserving (VP)-SDE\cite{score-based-SDE} linearly depicts the deterministic trend of energy reduction, signifying that the mean of the image signal deterministically decays to zero. The reverse SDE is derived from the forward one, leveraging the score function of the marginal probability densities. Image reconstruction is accomplished through the reverse process based on the reverse SDE, employing acquired $k$-space data as guidance. Notably, score-based MR reconstruction learns the data distribution rather than an end-to-end mapping between $k$-space data and images, enabling unsupervised learning and facilitating adaptation to out-of-distribution data. The method has demonstrated promising results, as reported in the literature.

Diffusion models have proven successful in MRI. Many current diffusion model-based reconstruction methods are primarily designed in the image domain, relying on spatial sensitivity maps of individual coils as weighting functions for multi-coil images. However, accurately measuring coil sensitivity maps (CSMs) can be challenging, especially when the field of view (FOV) is limited \cite{uecker2014eSPIRiT} or when singular points of phase exist \cite{blaimer2016comparison,uecker2017estimating}. Even minor estimation errors in coil sensitivities can lead to inconsistency between coil images and acquired $k$-space data, resulting in artifacts in the reconstructed image. On the other hand, direct interpolation of missing $k$-space data can circumvent the challenges associated with CSM estimation, as seen in parallel imaging (PI) methods in $k$-space, including GRAPPA\cite{grappa2002}, SPIRiT\cite{lustig2010spirit}, etc. These methods describe redundancy priors among multi-channel $k$-space data by estimating shift-invariant interpolation kernels, allowing for the estimation of missing $k$-space data with reduced sensitivity to inaccurate CSM estimation. As a result, $k$-space interpolation methods exhibit greater robustness compared to image-domain methods.
Inspired by this, diffusion models relying on $k$-space interpolation may inherit the advantage of robustness to inaccurate CSM estimation. Furthermore, this approach integrates both channel redundancy and data distribution priors, improving the accuracy in interpolating missing $k$-space data at high accelerations\cite{liang2020deep}. However, current diffusion models are primarily formulated within the image domain and cannot be readily applied in this scenario.


To address this issue, we reevaluate the traditional $k$-space interpolation SPIRiT model from an optimization perspective. Viewing the iterative algorithm for solving the SPIRiT model as the discrete Euler form of certain SDEs, we draw inspiration from this perspective to propose a novel diffusion-based MR reconstruction method. In this method, the self-consistent term of SPIRiT plays a crucial role as the primary drift coefficient in the SDE. Additionally, the CSM is introduced into the diffusion coefficient of this SDE to accurately calculate the mean and variance of the perturbation kernel. While our proposed method still necessitates the incorporation of CSM, its primary focus is on multi-channel $k$-space interpolation rather than single-channel image reconstruction synthesized based on CSM. As a result, our proposed model exhibits robustness to inaccurate sensitivity estimation. Since this method is inspired by SPIRiT in terms of self-consistency, we refer to it as SPIRiT-Diffusion.


\subsection{Contributions}
\begin{enumerate}
    \item To bridge the existing gap between image-domain diffusion models and $k$-space interpolation, we propose a new diffusion model for $k$-space interpolation with the SDE derived from the self-consistency prior of $k$-space data, specifically inspired by SPIRiT. This effectively mitigates the impact of inaccurate CSM estimation on image-domain diffusion models.

    \item Methodologically, we introduce a novel paradigm for SDE design in the diffusion model, termed model-driven diffusion. This entails designing an SDE that remains consistent with the physical significance of a given optimization model. That is, the drift term and diffusion coefficient of the SDE can be intricately designed based on the iterative solution of an optimization model.

\end{enumerate}

This paper is organized as follows. Section \ref{Background} introduces the background; Section \ref{Methods} presents the proposed method and implementation details; Section \ref{experiments} provides the experimental results. The discussion and conclusion are in Sections \ref{discussion} and \ref{conclusions}, respectively. 

\section{Background}\label{Background}

\subsection{SPIRiT Method}

The forward model for multi-channel MR reconstruction can be expressed as follows:
\begin{equation}
    \mathbf{y}=\mathbf{A}\mathbf{x}+\mathbf{n},
    \label{MR forward model}
\end{equation}
where $\mathbf{x}:=[\mathbf{x}_1,...,\mathbf{x}_m]$ is the image to be reconstructed and $\mathbf{x}_i$ is the $i$th channel image, $\mathbf{y}$ is the undersampled $k$-space data, $\mathbf{n}$ is the Gaussian noise, $\mathbf{A}$ is the encoding matrix with $\mathbf{A}:=\mathbf{M}\mathbb{F}$, $\mathbf{M}$ is the undersampling operator and $\mathbb{F}$ denotes Fourier transform.

Reconstructing a multi-channel MR image is akin to interpolating missing data from undersampled $k$-space data $\mathbf{y}$. Achieving precise interpolation inevitably requires the utilization of prior information from the $k$-space data. In this context, SPIRiT serves as a model that exploits the self-consistency prior in $k$-space. The self-consistency prior embodies a statistical regularity in $k$-space, where any point can be linearly interpolated based on its local points, including itself. In other words, a self-interpolation kernel $\mathbf{G}$ exists, enabling the multi-channel $k$-space data $\mathbf{\hat{x}}$ to be interpolated as $\mathbf{G\hat{x}}=\mathbf{\hat{x}}$, where $\mathbf{\hat{x}}$ is the $k$-space data with $\mathbf{\hat{x}}=\mathbb{F}(\mathbf{x})$.
Therefore, the regularization model, utilizing the self-consistency prior to solve the inverse problem \eqnref{MR forward model}, can be formulated as follows:
\begin{equation}
    \min_\mathbf{\hat{x}} \|\mathbf{G\hat{x}}-\mathbf{\hat{x}}\|_2^2+\mu\|\mathbf{A}\mathbf{x}-\mathbf{y}\|_2^2,
    \label{SPIRiT optimization}
\end{equation}
where the first term represents the self-consistency constraint, the second term enforces consistency with the data acquisition, and $\mu$ is the regularization parameter. 
In reality, when the measured data $\mathbf{y}$ often includes noise disturbances, we do not strictly enforce $\mathbf{G\hat{x}}=\mathbf{\hat{x}}$. Typically, a trade-off between self-consistency and data consistency can be achieved by adjusting the parameter $\mu$. Additionally, the optimization process can be controlled by setting a termination criterion, allowing the process to conclude when self-consistency reaches a certain level of accuracy to prevent overfitting.

\subsection{Score-Based Diffusion Model}
Score-based diffusion model is a framework of diffusion generative models. It perturbs data by injecting different scales of Gaussian noise to gradually transform the data distribution to Gaussian distribution and then generates samples from the Gaussian noise according to the corresponding reverse process. 
The diffusion process $\{\mathbf{x}(t)\}_{t=0}^T$ can be seen as the solution of the forward SDE as follows:
\begin{equation}
    \mathrm{d} \mathbf{x}=\mathbf{f}(\mathbf{x}, t) \mathrm{d} t+g(t) \mathrm{d} \mathbf{w},
    \label{background: forward sde}
\end{equation}
where $t$ is the continuous time variable, $t \in [0, T]$, $\mathbf{x}(0) \sim p_0=p_{data}$, $\mathbf{x}(T) \sim p_T$ and $p_T$ is a prior distribution, typically using Gaussian distribution. $\mathbf{f}$ and $g$ are the drift and diffusion coefficients of $\mathbf{x}(t)$, and $\mathbf{w}$ is the standard Wiener process.

The reverse-time SDE of \eqnref{background: forward sde} is:
\begin{equation}
    \mathrm{d} \mathbf{x}=\left[\mathbf{f}(\mathbf{x}, t)-g(t)^{2} \nabla_{\mathbf{x}} \log p_{t}(\mathbf{x})\right] \mathrm{d} t+g(t) \mathrm{d} \mathbf{\bar w},
\end{equation}
where $\mathbf{\bar w}$ is the standard Wiener process for the time from $T$ to $0$. The score function $\nabla_{\mathbf{x}} \log p_{t}(\mathbf{x})$ is approximated by the score model $\mathbf{s}_{\boldsymbol{\theta}}$ trained by
\begin{multline}
        \boldsymbol{\theta}^{*}=\underset{\boldsymbol{\theta}}{\arg \min } \mathbb{E}_{t}\Big\{\lambda(t) \mathbb{E}_{\mathbf{x}(0)} \mathbb{E}_{\mathbf{x}(t) \mid \mathbf{x}(0)}\big[\big\|\mathbf{s}_{\boldsymbol{\theta}}(\mathbf{x}(t), t)\\-\nabla_{\mathbf{x}(t)} \log p_{0 t}(\mathbf{x}(t) \mid \mathbf{x}(0))\big\|_{2}^{2}\big]\Big\},
    \label{SDE-loss}
\end{multline}
where $p_{0 t}(\mathbf{x}(t) \mid \mathbf{x}(0))$ is the perturbation kernel and can be derived from the forward diffusion process, $\mathbb{E}_{t}$ refers to the expectation with respect to time $t$. Once the score model $\mathbf{s}_{\boldsymbol{\theta}}$ is trained, we can generate samples through reverse-time SDE.

Score-based diffusion model has been applied to MR reconstruction and has shown promising results. Jalal et al.\cite{jalal2021robust} were the first ones to employ this method. They trained the score function using the fastMRI dataset and subsequently employed posterior sampling with Langevin to generate high-quality reconstructed data. The controllable generation is achieved through the reverse diffusion process with the posterior distribution $\nabla_{\mathbf{x}} \log p_t(\mathbf{x} \mid \mathbf{y})$:
\begin{equation}
    \begin{aligned}
        \nabla_{\mathbf{x}} \log p_{t}(\mathbf{x}(t) \mid \mathbf{y}) & = \nabla_{\mathbf{x}} \log p_{t}(\mathbf{x}(t)) + \nabla_{\mathbf{x}} \log p_{t}(\mathbf{y} \mid \mathbf{x}(t))\\
        & \approx \mathbf{s}_{{\boldsymbol{\theta}}^*}(\mathbf{x}(t), t) + \nabla_{\mathbf{x}} \log p_{t}(\mathbf{y} \mid \mathbf{x}(t))\\
        & = \mathbf{s}_{{\boldsymbol{\theta}}^*}(\mathbf{x}(t), t) + \frac{\mathbf{A}^H(\mathbf{y}-\mathbf{A} \mathbf{x}(t))}{\sigma^2_\epsilon},
    \end{aligned}
\end{equation}
where $\mathbf{y}$ is the undersampled $k$-space data and $\mathbf{x}$ is the desired MR image.

Using a similar approach, Song et al. utilized the score-based diffusion model to solve the inverse problems in both MRI and CT\cite{song2022solving}. However, they tested only on single-coil data. For multi-coil scenarios, Chung et al.\cite{chung2022score} proposed two approaches: one integrates the CSMs into the encoding matrix $\mathbf{A}$, while the other individually reconstructs coil images and then combines them using the sum-of-squares (SOS) method.

\section{Methodology}\label{Methods}
In this section, we will introduce the SPIRiT-Diffusion method, followed by a description of score model training and details on implementation.
\begin{figure*}[!t]
    \centerline{\includegraphics[width=1\textwidth]{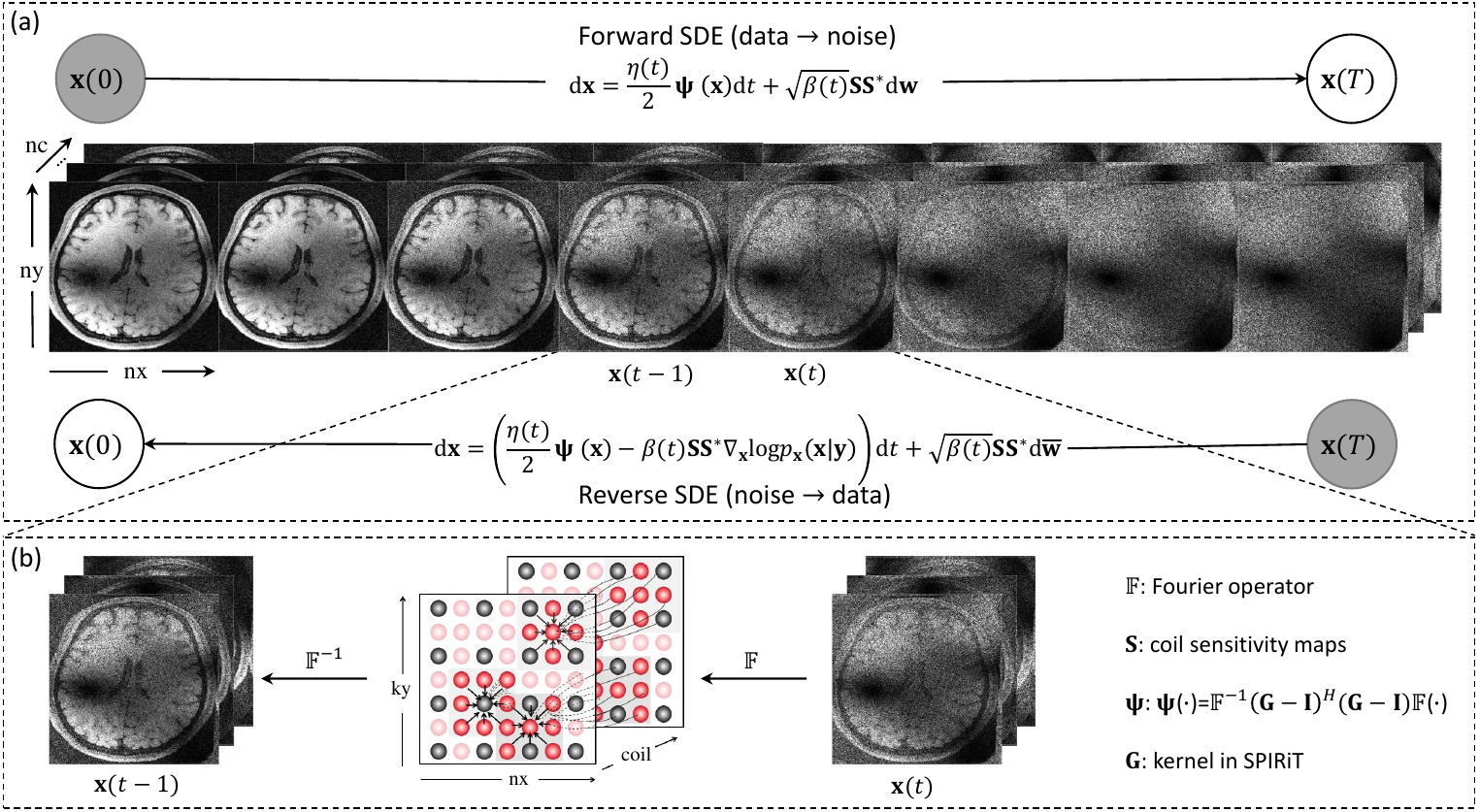}}
    \caption{The SPIRiT-Diffusion framework. (a) Forward process: Self-consistent noise at different scales is gradually injected into the multi-coil images. (b) Reverse process: From time $t$ to $t-1$, first, transform $\mathbf{x}(t)$ to the $k$-space domain, execute one iteration about the self-consistency term and the learned probability density prior term (i.e., score function), and finally transform the $k$-space data back to the image domain. By iteratively executing the reverse process, missing data in $k$-space are gradually filled in. It is important to note that the image related to self-interpolation of $k$-space data is referenced from \cite{lustig2010spirit}.
    }
    \label{fig:general}
\end{figure*}

\begin{figure}[!t]
    \centering
    \resizebox{0.9\columnwidth}{!}{\includegraphics{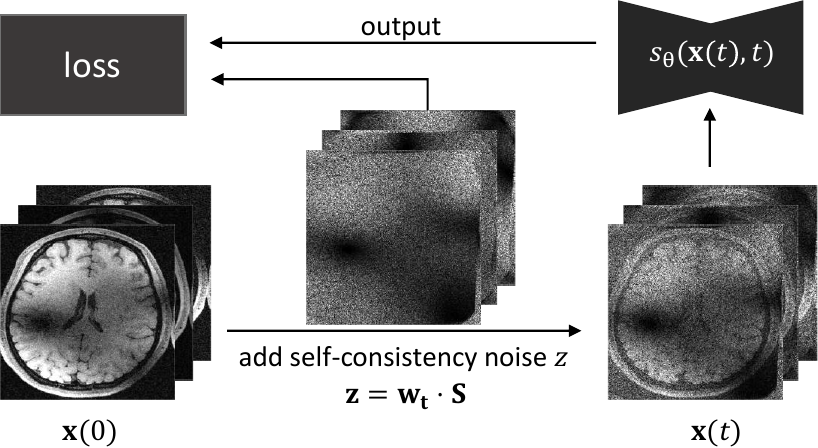}}
    \caption{Training Flowchart: Generate $\mathbf{x}(t)$ by adding self-consistent noise $\mathbf{z}$ to $\mathbf{x}(0)$ through a forward SDE, where $\mathbf{w}_t$ represents random noise at time $t$ and $\mathbf{S}$ denotes coil sensitivity. Feed $\mathbf{x}(t)$ into the network $\mathbf{s}_{\boldsymbol{\theta}}$ with $\mathbf{z}$ as the label, and use the network's output in conjunction with $\mathbf{z}$ to compute the training loss \eqnref{estimate score}.}
    \label{fig:training}
\end{figure}

\subsection{SPIRiT-Diffusion Model} 
\subsubsection{Forward SDE}
\eqnref{SPIRiT optimization} in the SPIRiT method can be solved using the stochastic gradient Langevin dynamics with the following iterative solution:
\begin{multline}
    \mathbf{x}_{k+1}=\mathbf{x}_{k}- \eta_{k}\mathbb{F}^{-1}((\mathbf{G}-\mathbf{I})^{H}(\mathbf{G}-\mathbf{I}) \mathbb{F}(\mathbf{x}_{k}))+ \\ \lambda_{k}\mathbf{A}^H(\mathbf{Ax_{k+1}}-\mathbf{y})-\beta_{k}\mathbf{z}_k, \quad k = 0, \cdots, T-1,
    \label{SPIRiT iterating}
\end{multline}
where $\eta_{k}$ and $\lambda_k :=\mu\eta_{k}$ denote the stepsize, $\beta_{k}:=\sqrt{2\eta_{k}\epsilon^{-1}}$, and $\epsilon$ denotes the inverse temperature parameter, $\mathbf{z} \sim \mathcal{N}(\mathbf{0}, \mathbf{I})$ and $\mathbf{I}$ denotes the identity matrix. Suppose \eqnref{SPIRiT iterating} is roughly interpreted as the discrete form of the reverse diffusion process conditioned on data consistency $\mathbf{Ax}-\mathbf{y}=\mathbf{n}$. Eliminating the impact of data consistency and regarding the gradient of the self-consistency term as the drift coefficient $\mathbf{f}(\mathbf{x}, t)$ of the diffusion process, we derive a new SDE as follows:
\begin{equation}
    \mathrm{d} \mathbf{x}=\frac{\eta(t)}{2}\Psi(\mathbf{x}) \mathrm{d} t+\sqrt{\beta(t)}\mathrm{d} \mathbf{w},
    \label{init SPIRiT SDE}
\end{equation}
where $\Psi(\cdot) := \mathbb{F}^{-1}(\mathbf{G}-\mathbf{I})^{H}(\mathbf{G}-\mathbf{I})\mathbb{F}(\cdot)$, $\beta(t)$ is the parameter to control the noise level, and $\eta(t)$ is the serialization of $\eta_{k}$. The related continuousization derivation is detailed in Appendix. \ref{Appendix:B}. The perturbation kernel can be derived by Eqs. 5.50 and 5.51 in \cite{sarkka2019applied} as:
\begin{multline}
    p_{0 t}(\mathbf{x}(t)\mid \mathbf{x}(0)) = \mathcal{N}\big(\mathbf{x}(t) ; e^{\frac{1}{2} \int_{0}^{t} \eta(s)\Psi d s}\mathbf{x}(0), \\\frac{1}{2} \int_{0}^{t} \beta(\tau)e^{\int_{\tau}^{t} \eta(s) d s} d \tau e^{2\Psi}\big), \quad t \in[0,1],
\end{multline}
where $\mathcal{N}(\mathbf{x}(t) ;\boldsymbol{\mu},\boldsymbol{\Sigma})$ indicates that a vector-valued variable $\mathbf{x}(t)$ follows a multivariate normal distribution with mean $\boldsymbol{\mu}$ and covariance matrix $\boldsymbol{\Sigma}$.
Since $\mathbf{\Psi}(\mathbf{x}(0))=0$, the mean $\boldsymbol{\mu}$ can be calculated using Taylor expansion as follows:
\begin{equation}
    \begin{aligned}
        \boldsymbol{\mu}=&e^{\frac{1}{2} \int_{0}^{t} \eta(s)\Psi d s}\mathbf{x}(0)\\
        =&\Big(\mathbf{I} + \frac{1}{2} \int_{0}^{t} \eta(s)\Psi d s + \frac{1}{2!}(\frac{1}{2} \int_{0}^{t} \eta(s)\Psi d s)^2+\\&o((\frac{1}{2} \int_{0}^{t} \eta(s)\Psi d s)^3)\Big)\mathbf{x}(0)\\
        =&\mathbf{x}(0).
    \end{aligned}
\end{equation}
But for the covariance matrix $\boldsymbol{\Sigma}$, the noise does not have the property of self-consistency, and the Taylor expansion of $e^{2\Psi}$ is computationally intolerable and time-consuming. We have tried using the first-order Taylor series as an approximation. However, the gap between this approximation and the real one is too large to learn the true data distribution, and experiments have shown poor performance.

To solve this issue, we introduce the coil redundancy to the diffusion coefficient, enforcing the self-consistency property to the Gaussian noise during the diffusion process, i.e. $\mathbf{GS}\mathbf{S}^*(\mathbf{z})=\mathbf{z}$ which implies $\mathbf{\Psi}(\mathbf{S}\mathbf{S}^*\mathbf{z})=0$, where $\mathbf{z} \sim \mathcal{N}(\mathbf{0}, \mathbf{I})$ and $\mathbf{S}$ is the CSM. The corresponding forward SPIRiT-Diffusion process is:
\begin{equation}
    \mathrm{d} \mathbf{x}=\frac{\eta(t)}{2}\Psi(\mathbf{x}) \mathrm{d} t+\sqrt{\beta(t)} \mathbf{S}\mathbf{S}^* \mathrm{d} \mathbf{w}.
    \label{forward SPIRiT}
\end{equation}
Then the perturbation kernel becomes:
\begin{multline}
    p_{0 t}(\mathbf{x}(t)\mid \mathbf{x}(0)) = \mathcal{N}\big(\mathbf{x}(t) ; \mathbf{x}(0), \\\frac{1}{2} \int_{0}^{t} \beta(\tau)e^{\int_{\tau}^{t} \eta(s) d s} d \tau e^{\Psi}\mathbf{S}\mathbf{S}^*[e^{\Psi}\mathbf{S}\mathbf{S}^*]^*\big), \quad t \in[0,1].
\end{multline}
Let $\sigma = \sqrt{\frac{1}{2} \int_{0}^{t} \beta(\tau)e^{\int_{\tau}^{t} \eta(s) d s} d \tau}$, the covariance can be calculated as follows:
\begin{equation}
    \begin{aligned}
        \boldsymbol{\Sigma^{\frac{1}{2}}}=&\sigma e^{\Psi}\mathbf{S}\mathbf{S}^*(\cdot)\\
        =&\sigma (\mathbf{I}+\Psi + \frac{1}{2!}\Psi^2 + o(\Psi^3))\mathbf{S}\mathbf{S}^*(\cdot)\\
        =&\sigma \mathbf{S}\mathbf{S}^*(\cdot).
    \end{aligned}
\end{equation}
The final expression of the perturbation kernel is
\begin{equation}
    p_{0 t}(\mathbf{x}(t)\mid \mathbf{x}(0)) = \mathcal{N}\big(\mathbf{x}(t); \mathbf{x}(0), \sigma^2 \mathbf{S}\mathbf{S}^*\big), \quad t \in[0,1].
    \label{perturbation kernel}
\end{equation}

\subsubsection{Estimating Score Functions}
In the training process,  \( \mathbf{x}(t) \) is obtained by adding noise to \( \mathbf{x}(0) \) in the forward process. Then, using \( \mathbf{x}(t) \) as the input for \( S_\theta \), the corresponding added noise serves as a training label for the model, training \( S_\theta \) as an approximation for the score function \( \nabla_\mathbf{x} \log p_t(\mathbf{x}) \). Since CSM has been integrated into the diffusion coefficient of the SPIRiT-Diffusion, it has become infeasible to train the score model by estimating pure Gaussian noise. According to the perturbation kernel in \eqnref{perturbation kernel}, the loss function of the score model in  SPIRiT-Diffusion is derived as (details in Appendix. \ref{Appendix:A}):
\begin{multline}
        \boldsymbol{\theta}^{*}=\underset{\boldsymbol{\theta}}{\arg \min } \mathbb{E}_{t}\big\{\lambda(t) \mathbb{E}_{\mathbf{x}(0)} \mathbb{E}_{\mathbf{x}(t) \mid \mathbf{x}(0)}\big[\big\|\sigma \mathbf{S}^*\mathbf{s}_{\boldsymbol{\theta}}(\mathbf{x}(t), t)\\+\mathbf{S}^*\mathbf{z}\big\|_{2}^{2}\big]\big\}.
        \label{estimate score}
\end{multline}

\subsubsection{Reverse SDE}
After the score model has been trained, MR images can be reconstructed by the conditional reverse SPIRiT-Diffusion process as:
\begin{multline}
    \mathrm{d} \mathbf{x}=\Big[\frac{\eta(t)}{2}\Psi(\mathbf{x}) -\beta(t) \mathbf{S}\mathbf{S}^*\nabla_{\mathbf{x}} \log p_{t}(\mathbf{x} \mid \mathbf{y})\Big]\mathrm{d} t \\ +\sqrt{\beta(t)}  \mathbf{S}\mathbf{S}^* \mathrm{d} \mathbf{\bar w}.
\end{multline}
The overall framework of SPIRiT-Diffusion is depicted in Fig. \ref{fig:general}. In the training process, $\mathbf{x}(0)$ is augmented with self-consistency noise $\mathbf{z}$ to obtain $\mathbf{x}(t)$, as illustrated in Fig. \ref{fig:training}. The network is trained with $\mathbf{x}(t)$ as input and $\mathbf{z}$ as the label. For the reconstruction process, the Predictor-Corrector method (PC Sampling) is employed to generate magnetic resonance images, as outlined in \algoref{alg:PC Sampling}.

\begin{algorithm}
	\caption{PC Sampling (SPIRiT-Diffusion).}
	\label{alg:PC Sampling}
	\begin{algorithmic}[1]
	    \Require{$\eta(t), \{\beta_i\}_{i=1}^N, \sigma, \mathbf{S}, \mathbf{y}, \lambda_1, \lambda_2, r, N, \mathbf{M}$.}
	   \State{$\mathbf{x}_{N} \sim \mathcal{N}(\mathbf{0}, (\sigma\mathbf{S}\mathbf{S}^*)^2)$}
	    \For{$i = N-1$ to $0$}
	        \State{$\mathbf{z} \sim \mathcal{N}(\mathbf{0}, \mathbf{I})$}
	        \State{$\mathbf{g} \leftarrow \mathbf{s}_{\boldsymbol{\theta^*}}\left(\mathbf{x}_{i+1}, i+1\right)$}
	        \State{$\mathbf{m}=\sum_{j=1}^{n} \mathbf{S}^{*} \mathbb{F}^{-1}\left(\mathbb{F}(\mathbf{S} \cdot \mathbf{x}_i) \cdot \mathbf{M} - \mathbf{y}\right)$}
	        \State{$\epsilon \leftarrow \lambda_1\left(\|\mathbf{g}\|_{2} /\|\mathbf{m}\|_{2}\right)$}
	        \State{$\mathbf{x}_{i} \leftarrow \mathbf{x}_{i+1}+\frac{1}{2}\eta_{i+1}\Psi(\mathbf{x}_i)+\beta_{i+1}\mathbf{SS}^*(\mathbf{g}-\epsilon\mathbf{m})+\sqrt{\beta_{i+1}}\mathbf{SS}^*(\mathbf{z})$}
            \For{$k = 1$ to $M$}
                \State{$\mathbf{z} \sim \mathcal{N}(\mathbf{0}, \mathbf{I})$}
                \State{$\mathbf{g} \leftarrow \mathbf{s}_{\boldsymbol{\theta}^*}\left(\mathbf{x}_{i}^{k-1}, i\right)$}
                \State{$\mathbf{m}=\sum_{j=1}^{n} \mathbf{S}^{*} \cdot \mathbb{F}^{-1}\left(\mathbb{F}(\mathbf{S} \cdot \mathbf{x}_i^k) \mathbf{M} - \mathbf{y}\right)$}
                \State{$\epsilon_1 \leftarrow 2 \left(r\|\mathbf{z}\|_{2} /\|\mathbf{g}\|_{2}\right)^{2}$}
                \State{$\epsilon_2 \leftarrow \lambda_2\left(\|\mathbf{g}\|_{2} /\|\mathbf{m}\|_{2}\right)$}
                \State{$\mathbf{x}_{i}^{k} \leftarrow \mathbf{x}_{i}^{k-1}+\frac{1}{2}\eta_{i+1}\Psi(\mathbf{x}_i^{k-1})+\epsilon_1 \mathbf{SS}^*(\mathbf{g}-\epsilon_2\mathbf{m})+\sqrt{2 \epsilon_1} \mathbf{SS}^*(\mathbf{z})$}
            \EndFor
            \State{$\mathbf{x}_{i-1}^{0} \leftarrow \mathbf{x}_{i}^{M}$}
        \EndFor
        \item[]
        \Return{${\mathbf{x}}_0^0$}
	\end{algorithmic}
\end{algorithm}

\subsection{Implementation Details}
The network structure of SPIRiT-Diffusion is the same as that of VE-SDE (\texttt{ncsnpp}\footnote{\url{https://github.com/yang-song/score_sde_pytorch}\label{sde code}}). The exponential moving average (EMA) rate is set to $0.999$, $\beta_{max}=348$, $\beta_{min}=0.01$, and the batch size is set to 1. Unlike previous approaches that combine multi-coil data into a single channel for network training, SPIRiT-Diffusion directly inputs multi-coil images to the network. The complex MR data is split into real and imaginary components and concatenated before input into the network, resulting in an input tensor of size $nc\times 2 \times nx \times ny$. $nc$ is the number of the coils, $2$ represents the concatenated real and imaginary parts of the data, and $nx$ and $ny$ represent the image size. The coil dimension is permuted to the batch size dimension to keep the convolution parameters of each channel consistent. The network is trained for 500 epochs in a computing environment using the PyTorch 1.13 library\cite{paszke2019pytorch}, CUDA 11.6 on an NVIDIA A100 Tensor Core GPU.

\section{Experiments}\label{experiments}
\subsection{Experimental Setup}
\subsubsection{Datasets}
We conducted experiments on a 3D joint Intracranial and Carotid Vessel Wall Imaging (VWI) dataset, including fully sample k-space data from 13 healthy volunteers and prospectively undersampled data from 4 stroke patients with acceleration factor (R) of 4.5. The dataset was acquired using the variable flip angle fast 3D spin-echo sequence on a 3T MR scanner (uMR 790, United Imaging Healthcare, China) with the combination of a 32-channel head coil and an 8-channel neck coil. The imaging parameters were: acquisition matrix $= 384 \times 318 \times 240$-$256$ $(RO \times PE1 \times PE2)$, FOV $= 232 \times 192 \times 144$-$154$ $(HF \times AP \times LR)$ mm$^3$, acquisition resolution $= 0.6$ mm$^3$, TR/TE $= 800/135$ ms, echo train length $= 46$, and elliptical scanning is used. The scan time of full sample $= 20$ to $22$ minutes. The inverse Fourier transform was initially applied in the RO direction to facilitate network operations, followed by splitting the 3D $k$-space data into 2D slices along the RO direction. Subsequently, coil compression \cite{zhang2013coil} was employed to compress the data to 18 channels. Ten healthy volunteers were randomly selected as the training dataset, with a total of 3680 images. The remaining data from the healthy volunteers was used as a test dataset \uppercase\expandafter{\romannumeral1} (768 images), and the data from the stroke patients was used as a test dataset \uppercase\expandafter{\romannumeral2} (968 images). To facilitate network operations, zero-padding of $k$-space data was performed in PE directions to increase the image size to $320 \times 320$. During testing, the padded k-space region was removed from the network output to obtain the final images, ensuring a consistent size with the ground truth obtained from the fully sampled data. We estimated the sensitivity maps using ESPIRiT \cite{uecker2014eSPIRiT} implemented in the BART toolbox \cite{uecker2016bart} with a $48\times48$ $k$-space center region.

\subsubsection{Parameter Configuration}
SPIRiT-Diffusion is compared with different types of reconstruction methods, including SPIRiT\cite{lustig2010spirit}, DL-SPIRiT\cite{jiadeep}, ISTA-Net\cite{zhang2018ista}, DDS\cite{chung2024decomposed}, VE- and VP-SDE based on CSMs\cite{chung2022score}. The deep learning-based methods were trained using identical training data. In ISTA-Net, the learning rate is set to 0.0001 with a batch size of 4, while in DL-SPIRiT, the learning rate is set to 0.001 with a batch size of 5. ISTA-Net underwent 50 epochs of training, while DL-SPIRiT underwent 40 epochs. All score-based SDE methods underwent 500 epochs of training, with the noise scale $N=1000$. The EMA rate was 0.9999 for VP-SDE and 0.999 for VE-SDE, DDS. VP-SDE controlled the noise level during forward diffusion by setting $\beta_{max}=20$, and $\beta_{min}=0.1$  while VE-SDE controlled the noise level in forward diffusion by setting $\sigma_{max}=348$ and $\sigma_{min}=0.1$\cite{song2020improved}. The network used in  VE-SDE, DDS, and VP-SDE was the U-net architecture improved by Song et al.(\textit{i.e.}, \texttt{ncsnpp}\footnote{\url{https://github.com/yang-song/score_sde_pytorch}\label{sde code}} in the code of score-based SDEs).

 \subsubsection{Performance Evaluation}
Three metrics were used to evaluate the results quantitatively, including normalized mean square error (NMSE), the peak signal-to-noise ratio (PSNR), and the structural similarity index (SSIM)\cite{SSIM}. The metrics were calculated on the imaging region, excluding the background, which was determined by threshold segmentation with a threshold set at 10\% of the maximum image intensity.  

\subsection{Experimental Results}
\subsubsection{Retrospective Experiments}\label{Retrospective}
\begin{figure*}[!t]
    \centerline{\includegraphics[width=1\textwidth]{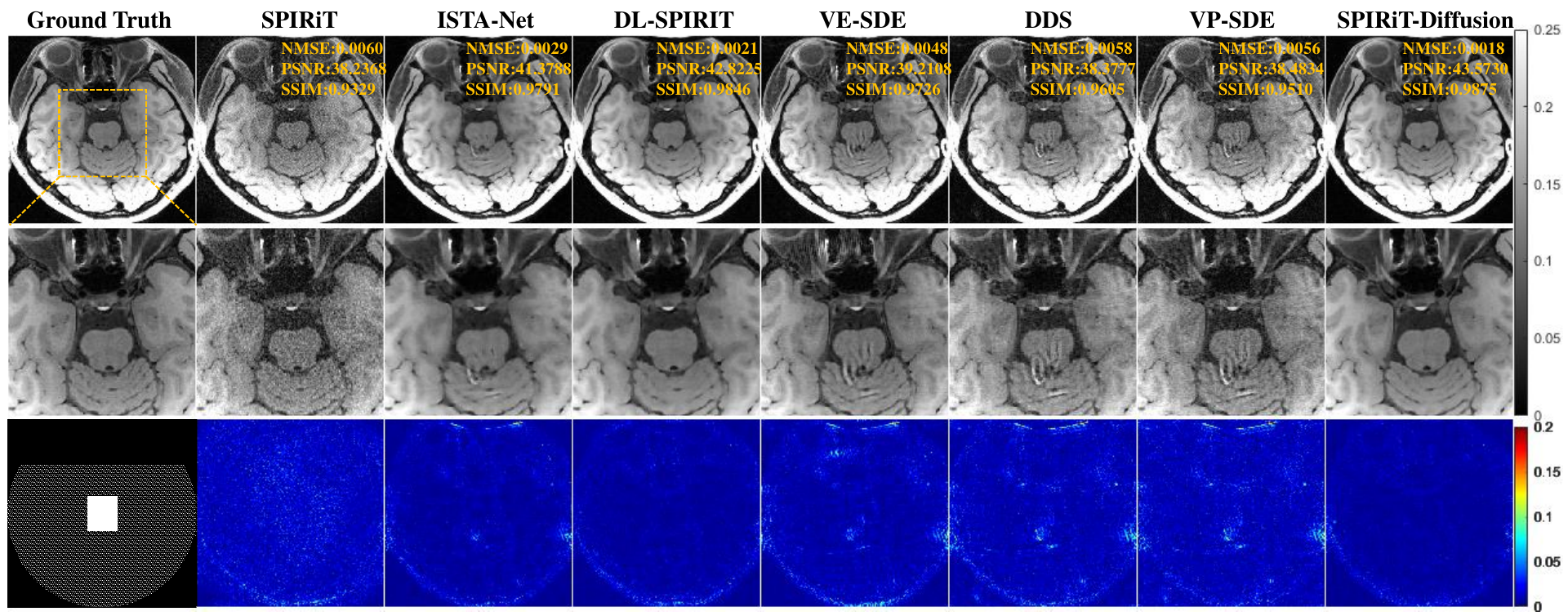}}
    \caption{Reconstruction results of VWI data at absolutely R = 7.6 The top row shows the ground truth and the reconstructions obtained using different methods. The second row shows an enlarged view of the ROI, and the third row displays the error map of the reconstructions. }
    \label{fig: retrospective}
\end{figure*}


\begin{figure*}[!t]
    \centerline{\includegraphics[width=1\textwidth]{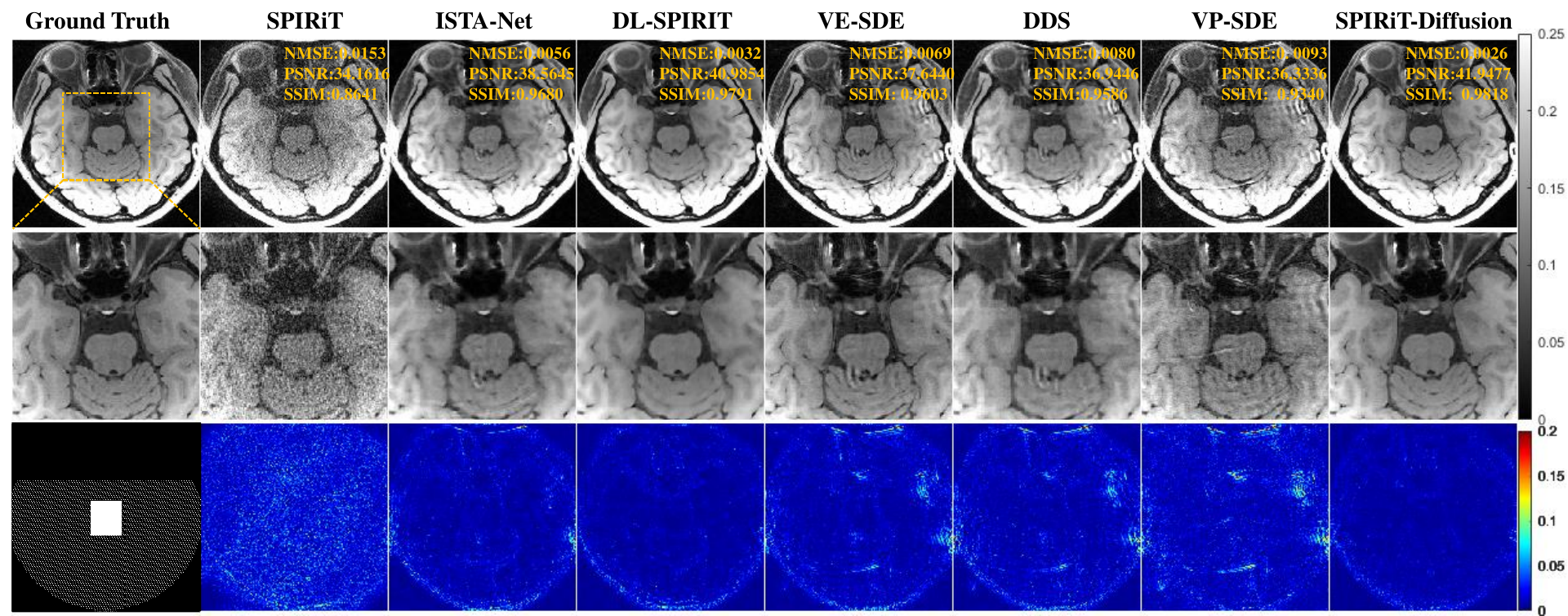}}
    \caption{Reconstruction results of VWI data at R = 10. The top row shows the ground truth and the reconstructions obtained using different methods. The second row shows an enlarged view of the ROI, and the third row displays the error map of the reconstructions.}
    \label{fig: acc10}
\end{figure*}
\begin{figure*}[!t]
    \centerline{\includegraphics[width=1.0\textwidth]{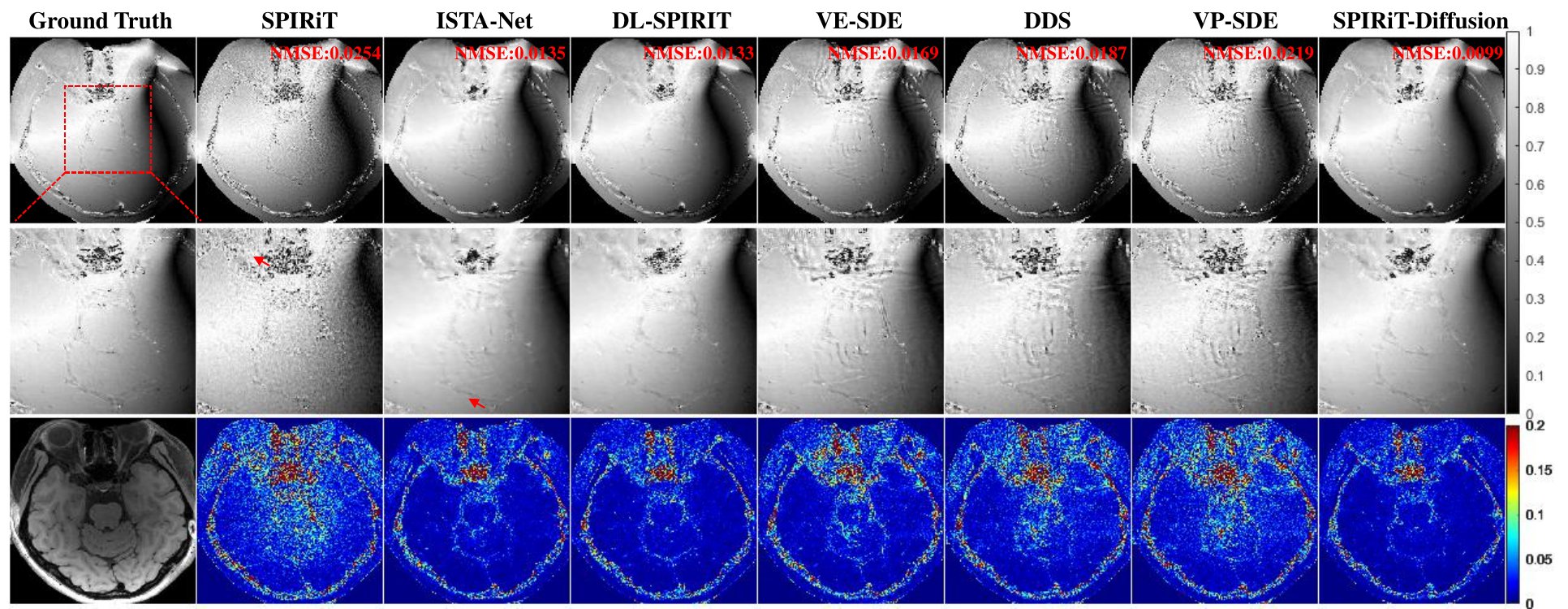}}
    \caption{Phase reconstruction results of VWI data at  R= 7.6. The top row shows the ground truth and the reconstructions obtained using different methods. The second row shows an enlarged view of the ROI, and the third row displays the error map of the reconstructions.}
    \label{fig: phase_csm76}
\end{figure*}

\begin{figure*}[!t]
    \centerline{\includegraphics[width=1.0\textwidth]{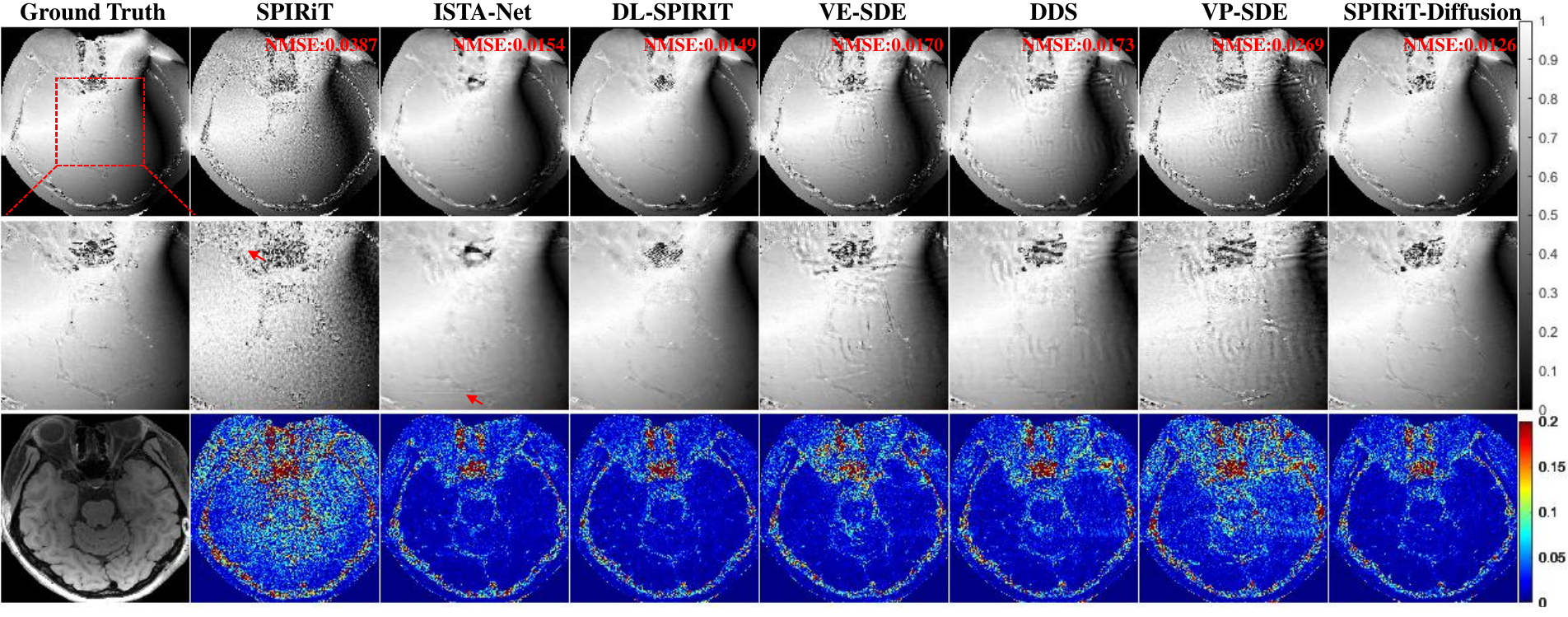}}
    \caption{Phase reconstruction results of VWI data at R = 10. The top row shows the ground truth and the reconstructions obtained using different methods. The second row shows an enlarged view of the ROI, and the third row displays the error map of the reconstructions.}
    \label{fig: phase_csm}
\end{figure*}

We retrospectively undersampled the test dataset \uppercase\expandafter{\romannumeral1} with R = 7.6 and 10, respectively. \figref{fig: retrospective} shows the reconstruction results of R = 7.6. The result of SPIRiT shows severe noise amplification since it is based on the parallel imaging method. DL-SPIRiT can suppress noise but also lose tiny structures, which is a common issue for DL-based reconstructions. The reconstruction quality of the SDEs method, except SPIRiT-Diffusion, is poor, and artifacts can be easily seen in the image. This is because the estimated CSMs are inaccurate due to the FOV aliasing, especially at the image’s border. SPIRiT-Diffusion achieves good reconstruction quality with the highest quantitative metrics among compared methods.

\figref{fig: acc10} shows the results of R = 10. SPIRiT-Diffusion can still achieve good reconstruction quality with quantitative metrics slightly degraded. The vessel wall can be barely seen in the images using other compared reconstruction methods at such a high acceleration rate. More aliasing artifacts appear on the images reconstructed by VE-SDE, DDS, VP-SDE, and ISTA-Net relative to R = 7.6. 

\figref{fig: phase_csm76} and \figref{fig: phase_csm}  show the phase  images of the reconstructed VWI data with R = 7.6 and 10, respectively. ISTA-Net and DL-SPIRiT demonstrate superior phase reconstruction performance compared to other methods. This superiority is primarily attributed to the fact that typical phase images are smooth, and the loss of details doesn’t significantly impact image quality. Diffusion-based reconstructions, including DDS, VE-, and VP-SDE, show notable phase distortions due to aliasing artifacts. SPIRiT-Diffusion demonstrates visual quality similar to that of ISTA-Net and DL-SPIRiT but with more preserved details. 

Table. \ref{tab: retrospective} shows the average quantitative metrics across all slices of test dataset \uppercase\expandafter{\romannumeral1}. SPIRiT-Diffusion achieved the best NMSE and SSIM.


\subsubsection{Prospective Experiments}\label{Prospective}
\begin{figure*}[!t]
    \centerline{\includegraphics[width=1\textwidth]{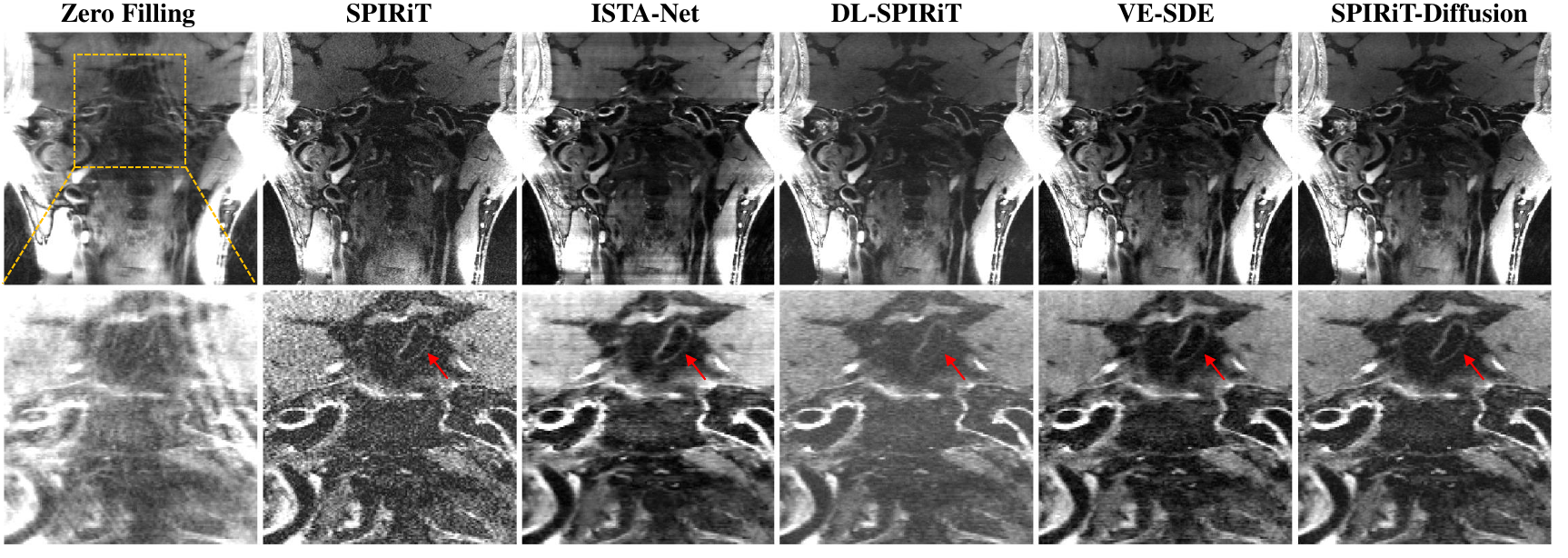}}
    \caption{
    Prospective reconstruction results of VWI data at R = 4.5, with images reformatted in the coronal direction. The top row displays the ground truth alongside reconstructions obtained using various methods. The second row presents an enlarged view of the region of interest (ROI) marked by the yellow box in the first row. The SPIRiT image appears somewhat noisy. Aliasing artifacts are visible in the images from ISTA-Net and VE-SDE. The DL-SPIRiT result is missing parts of the vessel wall. SPIRiT-Diffusion demonstrates a strong capability to suppress artifacts and noise.
    }
    \label{fig: Prospective}
\end{figure*}


\figref{fig: Prospective} shows the reconstruction results of the prospectively undersampled VWI data from a stroke patient at the coronal view. Similar conclusions can be drawn from the retrospective study. The image of SPIRiT is a bit noisy. Aliasing artifacts appear on the images of ISTA-Net and VE-SDE. Parts of the vessel wall were lost on the images of DL-SPIRiT. SPIRiT-Diffusion can suppress artifacts and noise well, and the vessel wall can be clearly seen in its image.



To accurately assess the image quality in our prospective experiments, two experienced radiologists (with 4 and 6  years of experience) were recruited to assess the image quality using a four-point score. The criteria for each score are as follows: Score 4 (Excellent): The image clearly displays the entire boundary of the vessel wall, showcasing complete details; Score 3 (Good): The image presents the contour of the vessel wall fairly well, although a small portion of the boundary is blurred or invisible; Score 2 (Fair): The image is reasonably capable of displaying the vessel wall, with part of the boundary visible, but with more than a quarter of the area appearing blurred or unrecognizable; Score 1 (Nondiagnostic): Most of the vessel wall boundaries are unclear and indistinguishable. The image volumes reconstructed using different methods were placed in random order and readers were blinded to the reconstruction methods. The two radiologists performed evaluations independently.
 The results of the scoring are shown in Table. \ref{tab:doctor_score}, where we can observe that the SPIRiT-Diffusion method surpasses other experimental approaches in terms of image quality.



\begin{table}
  \caption{\label{tab: retrospective} The average quantitative metrics on test dataset \uppercase\expandafter{\romannumeral1} at R = 7.6 and 10.}
  \centering
  \resizebox{\linewidth}{!}{
      \begin{tabular}{c|cccc}
        \hline \hline R & Methods & NMSE(*e-2) & PSNR (dB) & SSIM(*e-2) \\
        \hline \multirow{8}{*}{7.6}
        & SPIRiT & 1.06 $\pm$ 0.82 & 37.58 $\pm$ 4.96 & 93.89 $\pm$ 5.70 \\
      &ISTA-Net &0.66 $\pm$ 0.67 & 39.61 $\pm$ 4.60 & 97.54 $\pm$ 2.12 \\
      &DL-SPIRiT & 0.44 $\pm$ 0.36 & 41.25 $\pm$ 4.62 & 97.86 $\pm$ 1.78 \\
      &VE-SDE & 0.63 $\pm$ 0.32 & 39.43 $\pm$ 4.40 & 97.07 $\pm$ 2.30 \\
      &DDS & 0.66 $\pm$ 0.39 & 39.37 $\pm$ 4.45 & 96.29 $\pm$ 3.37 \\
      &VP-SDE & 0.74 $\pm$ 0.42 & 38.90 $\pm$ 4.57 & 95.55 $\pm$ 4.09 \\
      &SPIRiT-Diffusion & \textbf{0.42 $\pm$ 0.25} & \textbf{41.30 $\pm$ 4.21} & \textbf{98.15 $\pm$ 1.50} \\

         \hline \multirow{7}{*}{10}&SPIRiT & 1.80 $\pm$ 1.37 & 35.33 $\pm$ 5.09 & 91.29 $\pm$ 8.55 \\
  &ISTA-Net & 1.03 $\pm$ 0.96 & 37.66 $\pm$ 4.64 & 96.79 $\pm$ 2.60 \\
  &DL-SPIRiT & 0.64 $\pm$ 0.66 & \textbf{39.70 $\pm$ 4.59} & 97.31 $\pm$ 2.14 \\
  &VE-SDE & 0.89 $\pm$ 0.51 & 37.96 $\pm$ 4.38 & 96.24 $\pm$ 2.78 \\
  &DDS & 1.06 $\pm$ 1.13 & 37.67 $\pm$ 4.59 & 95.70 $\pm$ 3.64 \\
  &VP-SDE & 1.17 $\pm$ 0.74 & 36.87 $\pm$ 4.25 & 95.80 $\pm$ 3.20 \\
  &SPIRiT-Diffusion & \textbf{0.64 $\pm$ 0.43} & 39.56 $\pm$ 4.19 & \textbf{97.51 $\pm$ 1.89} \\
  \hline \hline
    \end{tabular}
  }
\end{table}

\begin{table}[!t]
  \caption{\label{tab:doctor_score} 
The average subjective scores of two radiologists for the prospective experiment.}
  \centering
  \resizebox{\columnwidth}{!}{%
    \begin{tabular}{c|ccccc}
      \hline \hline
      Method & Patient 1 & Patient 2 & Patient 3 & Patient 4 & Mean Score \\
      \hline
      SPIRiT & 2.00 & 2.00 & 1.00 & 2.00 & 1.68 \\
      ISTA-Net & 4.00 & 3.00 & 3.00 & 4.00 & 3.46 \\
      DL-SPIRiT & 4.00 & 3.50 & 3.00 & 4.00 & 3.60 \\
      VE-SDE & 3.00 & 2.50 & 2.50 & 3.50 & 2.85 \\
      SPIRiT-Diffusion & \textbf{4.00} & \textbf{4.00} & \textbf{4.00} & \textbf{4.00} & \textbf{4.00} \\
      \hline \hline
    \end{tabular}%
  }
\end{table}

\section{Discussion}\label{discussion}
This study has demonstrated that the self-consistency property of $k$-space data can be incorporated into an SDE to interpolate $k$-space in diffusion-based MR reconstruction. Based on this, the proposed SPIRiT-Diffusion method for coil-by-coil reconstruction avoids the issues induced by inaccurate coil sensitivity estimation. Since this method exploits both data distribution prior and the correlations among multi-coil data, it can reconstruct high-quality images with high acceleration (R = 10) in 3D imaging. Even in cases where reconstruction by image domain methods failed, such as ISTA-Net and VE-SDE, SPIRiT-Diffusion can still achieve a solution. We chose 3D VWI data for testing since 3D VWI has a large FOV coverage, and a reduced FOV is typically employed to shorten the scan time. In this scenario, it is difficult to estimate CSMs accurately, which better verifies that SPIRiT-Diffusion improves the robustness against inaccurate CSM estimation. In addition, we developed model-driven diffusion that integrates a traditional optimization model into the diffusion model, making the diffusion process conform more to the constraints than pure data-driven diffusion. 

\subsection{The Effect of CSM Estimation Methods}

\begin{table}[!t]
  \caption{\label{tab: training_csm} The average quantitative metrics of SPIRiT-Diffusion using different CSM estimation methods on test dataset \uppercase\expandafter{\romannumeral1} at R = 7.6.}
  \centering
  \resizebox{\linewidth}{!}{
      \begin{tabular}{c|cccc}
        \hline \hline Training & Testing & NMSE(*e-2) & PSNR (dB) & SSIM(*e-2) \\
        \hline ESPIRiT & ESPIRiT  & 0.42  $\pm$ 0.25 & 41.30 $\pm$ 4.21 & 98.15 $\pm$ 1.50 \\
        ESPIRiT & SOS & 0.57 $\pm$ 0.38 & 39.88 $\pm$ 4.38 & 97.41 $\pm$ 2.19 \\
        SOS & ESPIRiT  & 0.45 $\pm$ 0.31 & 41.09 $\pm$ 4.24 & 98.14 $\pm$ 1.50 \\
        SOS & SOS & 0.59 $\pm$ 0.37 & 39.67 $\pm$ 4.55 & 97.36 $\pm$ 2.22 \\
        \hline \hline
    \end{tabular}
  }
\end{table}

\begin{figure*}[!t]
    \centerline{\includegraphics[width=0.75\textwidth]{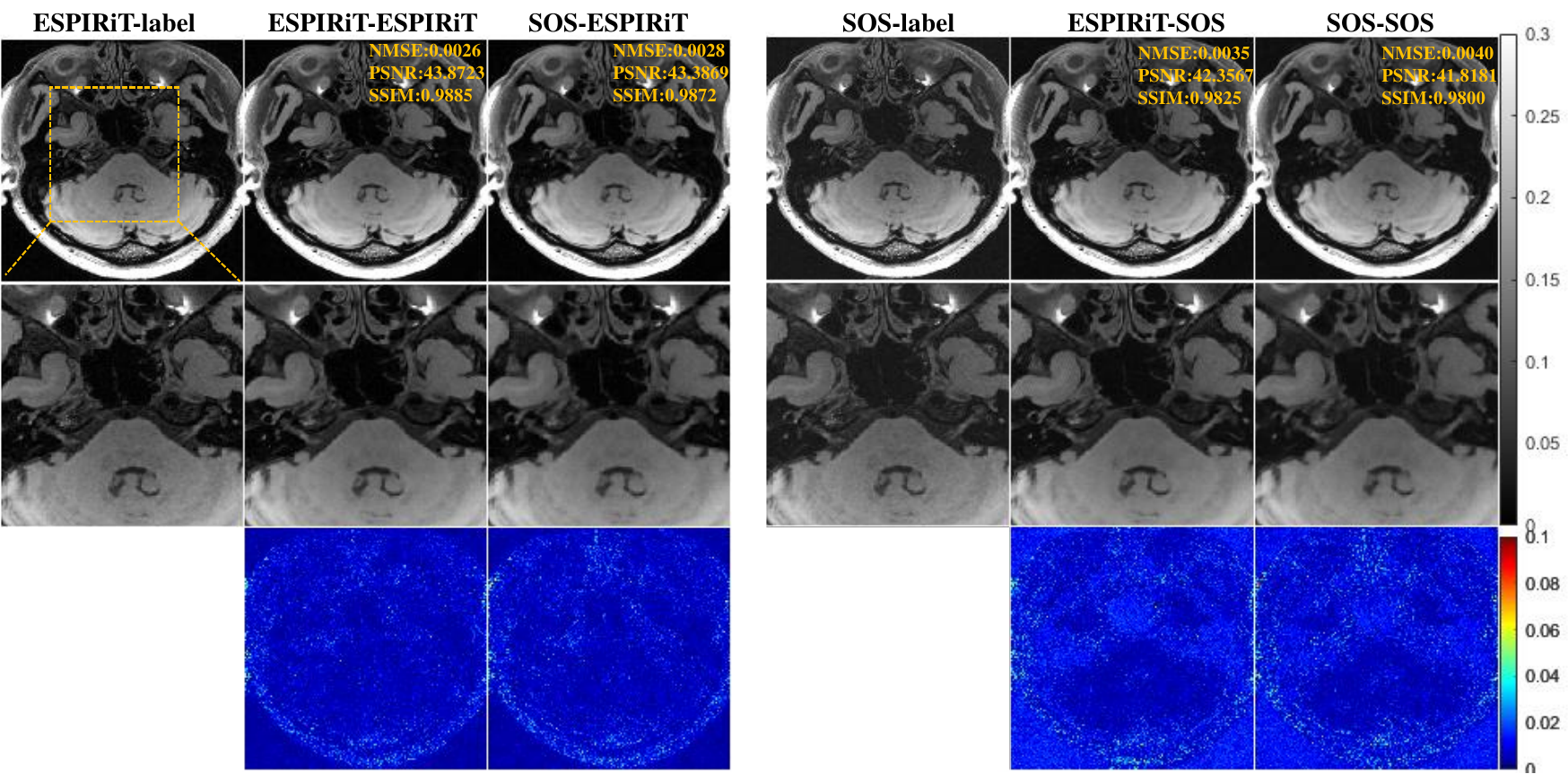}}
    \caption{SPIRiT-Diffusion Reconstruction results at R = 7.6 with different CSMs. 
    The term ``ESPIRiT-SO'' indicates the used CSM methods in training (ESPIRiT) and testing (SOS), respectively. Other terms are the same. The ESPIRiT/SOS-label denotes the combination of full-sampled k-space using CSMs of the ESPIRiT/SOS method. }
    \label{fig: diff_csm}
\end{figure*}


We conducted experiments to evaluate the impact of different CSM estimation methods on the reconstruction quality of SPIRiT-Diffusion. Two CSM methods were employed: one directly utilized the conjugate of low-resolution ($48\times48$ ACS region) coil images, as indicated in the SOS method; the other was the ESPIRiT method\cite{uecker2014eSPIRiT}, which is widely used in MR reconstruction research. The SPIRiT-Diffusion model was trained and tested using different CSMs to evaluate the influence on reconstruction quality. Please note that the CSM used in testing was also employed for coil combination to generate the final reconstruction. Therefore, the gold standard should utilize the corresponding CSM for a fair comparison. \figref{fig: diff_csm} shows the reconstructed results using SPIRiT-Diffusion at R = 7.6 using different types of CSM methods, and the corresponding quantitative results are depicted in Table. \ref{tab: training_csm}. Generally speaking, using either SOS or ESPIRiT methods in training can achieve similar results. The reconstruction performance using the SOS method in testing is slightly lower than those using the ESPIRiT method because of the low signal-to-noise ratio of SOS CSM in coil combination, but the visual difference between them is minimal. This experiment validates that although CSM is integrated as a weighting function in the perturbation kernel to reduce computational complexity, its variation has little effect on the performance of SPIRiT-Diffusion.

\begin{figure*}[!t]
    \centerline{\includegraphics[width=1\textwidth]{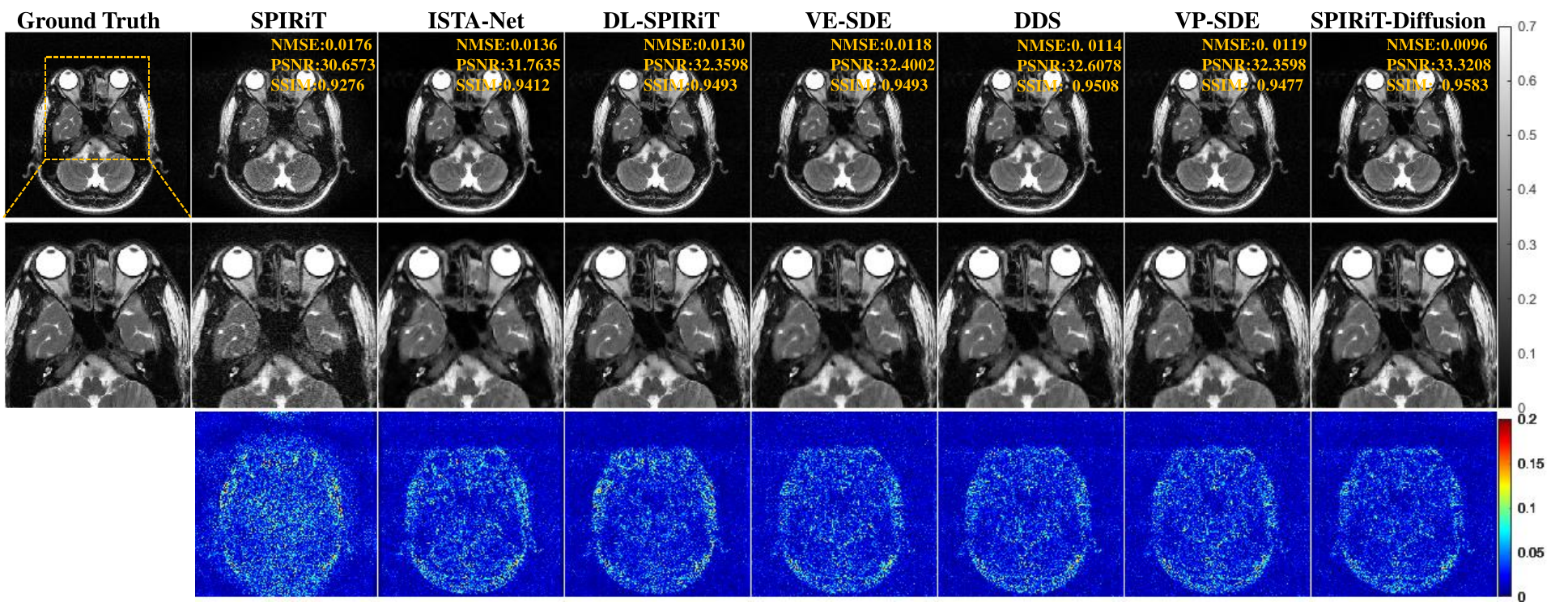}}
    \caption{Reconstruction results of brain T2w data at R = 7.6. SPIRiT-Diffusion achieves the best reconstruction quality.}
    \label{fig: fse}
\end{figure*}

\begin{figure*}[!t]
    \centerline{\includegraphics[width=1\textwidth]{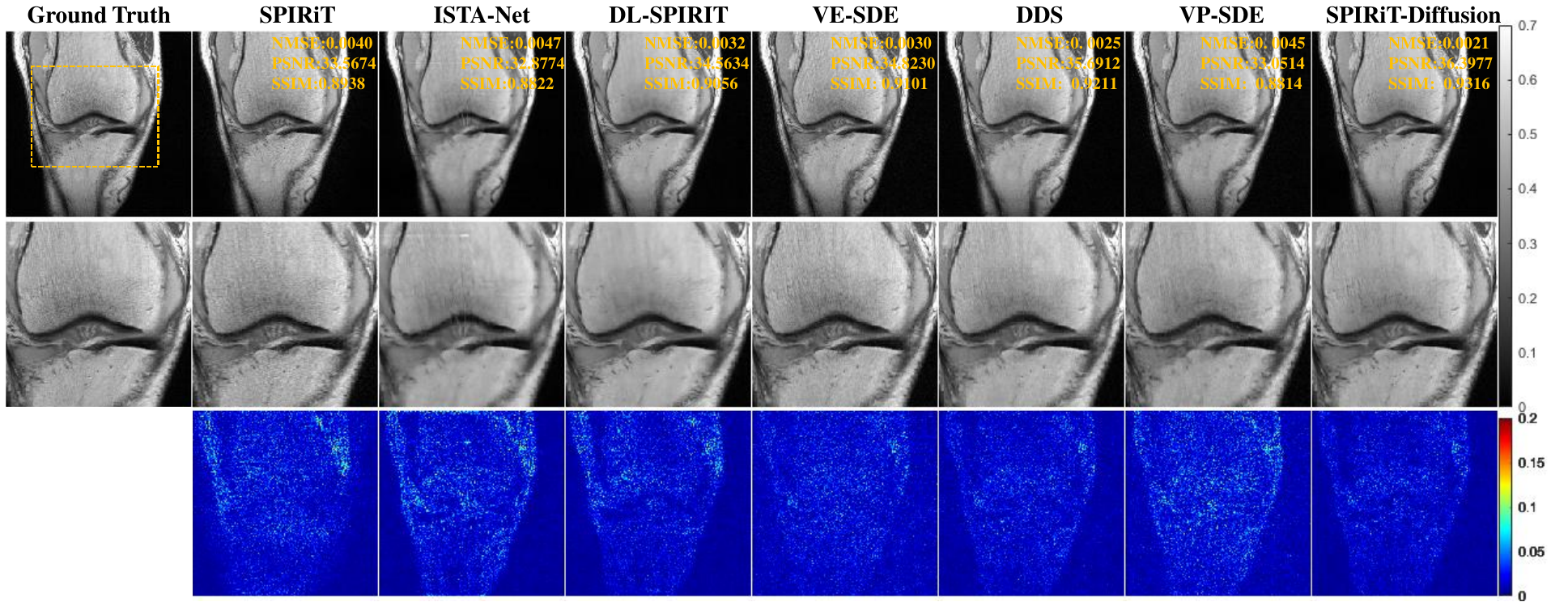}}
    \caption{Reconstruction results of fastMRI knee data at R = 7.6. SPIRiT-Diffusion achieves excellent reconstruction quality.}
    \label{fig: out_knee}
\end{figure*}

\subsection{The model-driven diffusion model}
Physics plays an important role in MR imaging and is generally utilized to construct the optimization model for MR reconstruction. However, conventional diffusion models are trained solely on image domain distributions and do not leverage these physical principles. To integrate imaging physics into diffusion models, we can roughly treat the iterative solution of an optimization model as the reverse diffusion process. That is, the iterative solution would correspond to the drift coefficient of the reverse-time SDE. In this scenario, the physics in any optimization model can be used to “drive” the diffusion process by designing the SDE accordingly, dubbed model-driven diffusion. Model-driven diffusion reconstruction methods then utilize both priors from the optimization model and the data distribution, improving reconstruction quality. SPIRiT-Diffusion can be regarded as a typical example of model-driven diffusion.

In practice, we usually need to know the perturbation kernel to estimate the score function in diffusion models. Generally speaking, the perturbation kernel is a Gaussian distribution with closed-form mean and variance when the drift coefficient is affine.  For model-driven diffusion, it’s substantial to make the perturbation kernel computable since the physical operator is in the exponential term most of the time. In SPIRiT-Diffusion, this issue is solved by introducing the coil sensitivity in the diffusion coefficient. Alternatively, using sliced score matching for model training also provides a feasible solution, i.e., simulating the SDE to sample from $p_{0 t}(\mathbf{x}(t) \mid \mathbf{x}(0))$ to bypass the computation of the perturbation kernel.

\subsection{The Distinction From Other Diffusion-based MR reconstructions}
Recently, a series of studies have successfully applied diffusion models to MRI. The innovation of these methods can be succinctly summarized in two aspects. Firstly, by ingeniously coupling data consistency terms to existing reverse diffusion processes, these methods guide the accurate generation of MRI images\cite{chung2022score,gungor2023adaptive,peng2022towards}. Such approaches are predominantly designed in the image domain. Secondly, efforts have been directed towards designing new diffusion processes to achieve more robust performance\cite{chung2024decomposed,mirza2023super,cao2024high,xie2022measurement}. Compared to these diffusion methods, the novelty of SPIRiT-Diffusion lies in the following two aspects:

Unlike the first category, SPIRiT-Diffusion not only includes data consistency terms but also integrates additional MRI physical priors—$k$-space self-consistency—into the diffusion model. This bridges the gap between image domain diffusion models and $k$-space interpolation. The model's SDE derives from a self-consistent prior of $k$-space data inspired by SPIRiT. Consequently, this approach effectively reduces the impact of inaccurate CSM estimation on image domain diffusion models.

Compared to the second category, our method introduces a novel SDE design paradigm termed model-driven diffusion. This entails designing an SDE to align with the physical significance of a given optimization model. In essence, the drift and diffusion coefficients of the SDE can be meticulously tailored based on the iterative solution of the optimization model. Guided by MRI physical priors, this approach to diffusion process design offers greater interpretability and robustness compared to existing methods.

\subsection{The Distinction From Other DL-Based $k$-Space Interpolation Methods}

There are several deep learning (DL) reconstruction methods based on $k$-space interpolation, such as RAKI \cite{akccakaya2019scan}, LORAKI \cite{kim2019loraki}, $k$-space DL \cite{han2019k}, and Deep-SLR \cite{pramanik2020deep}. RAKI is rooted in the GRAPPA method, where a neural network is trained using autocalibration signal (ACS) data to learn the $k$-space channel redundancy instead of linear interpolation. Similarly, LORAKI extends this approach based on the auto-calibrated LORAKS method \cite{7164018}. $k$-space DL and Deep-SLR, on the other hand, learn the $k$-space channel redundancy using additional training data rather than ACS data.

These methods primarily rely on the prior of $k$-space channel redundancy. The difference lies in the representation of this prior: while traditional $k$-space interpolation methods employ linear interpolation, $k$-space DL methods often use more complex neural networks. Although neural networks can more accurately capture the relationships between multiple $k$-space channels compared to linear kernels, they cannot surpass the prior of channel redundancy.

In contrast to the aforementioned methods, the SPIRiT-diffusion method proposed in this paper utilizes SPIRiT linear interpolation to represent the redundancy prior while also introducing a diffusion model to explore the data distribution as a complementary prior. This approach yields a more complete prior beyond the sole channel redundancy prior of $k$-space data. Therefore, in scenarios with relatively high acceleration rates, SPIRiT-Diffusion outperforms traditional linear interpolation methods and $k$-space DL methods. Experimental results validate the aforementioned conclusion.


\subsection{Out-of-Distribution}

We conducted experiments to assess the generalizability of the SPIRiT-Diffusion model, which was trained using the VWI T1-weighted dataset, to different contrasts and body parts. For contrast variation, fully sampled brain T2-weighted (T2w) data was utilized, collected using the fast spin-echo sequence on a 1.5T scanner (SuperMark, Anke China) with an 8-channel head coil. The imaging parameters were: $\text{TR/TE}=3000/100$ ms, $\text{FOV} =230 \times 230$ mm, slice thickness $= 5$ mm, matrix size $= 320 \times 320$. The data was retrospectively undersampled with $R = 7.6$. Fig. 9 displays the representative reconstruction results using different methods and Table. \ref{tab: out} (top row) shows the average quantitative metrics of the T2w data. Overall, the performance of all methods degrades when the trained models are directly applied to the T2w contrast. \figref{fig: fse} presents the results with R = 7.6 in T2w brain data. Aliasing artifacts appear on the image reconstructed by ISTA-Net, while SPIRiT-Diffusion achieves an acceptable image quality without obvious aliasing artifacts.

Three T1w knee datasets in the public fastMRI dataset\footnote{\url{https://fastmri.org/}}\cite{zbontar2018fastmri, knoll2020fastmri} were utilized to test the generalizability of SPIRiT-Diffusion to different body parts. The k-space data were retrospectively undersampled with $R = 7.6$. \figref{fig: out_knee} presents the reconstruction results of $R = 7.6$ and Table. \ref{tab: out} (bottom row) shows the average quantitative metrics of the knee data. We found that diffusion-based models, including SPIRiT-Diffusion, VE- and VP-SDEs, exhibit acceptable image quality with relatively high quantitative performance. The supervised methods, i.e. ISTA-Net and DL-SPIRiT have aliasing artifacts on the reconstructed images. These results demonstrate that SPIRiT-Diffusion has better generalizability than the supervised DL methods.


\begin{table}
  \caption{\label{tab: out} The average quantitative metrics on the T1w knee data and T2W brain data at R = 7.6.}
  \centering
  \resizebox{\linewidth}{!}{
      \begin{tabular}{c|cccc}
        \hline \hline   & Methods & NMSE(*e-2) & PSNR (dB) & SSIM(*e-2) \\
         \hline \multirow{7}{*}{brain} & SPIRiT & 1.90 $\pm$ 0.47 & 32.67 $\pm$ 2.10 & 92.48 $\pm$ 3.05 \\
      & ISTA-Net & 1.66 $\pm$ 0.56 & 33.34 $\pm$ 2.27 & 94.63 $\pm$ 2.02 \\
      & DL-SPIRiT & 1.33 $\pm$ 0.31 & 34.18 $\pm$ 1.93 & 95.15 $\pm$ 1.78 \\
      & VE-SDE & 1.34 $\pm$ 0.24 & 34.09 $\pm$ 1.51 & 94.57 $\pm$ 1.88 \\
      & DDS & 1.23 $\pm$ 0.25 & 34.48 $\pm$ 1.67 & 95.31 $\pm$ 1.62 \\
      & VP-SDE & 1.48 $\pm$ 0.42 & 33.76 $\pm$ 1.79 & 94.68 $\pm$ 1.87 \\
      & SPIRiT-Diffusion & \textbf{1.22 $\pm$ 0.40} & \textbf{34.66 $\pm$ 2.10} & \textbf{95.77 $\pm$ 1.47} \\

        \hline \multirow{7}{*}{knee} 
      & SPIRiT & 0.51 $\pm$ 0.21 & 34.30 $\pm$ 2.47 & 89.60 $\pm$ 4.24 \\
      & ISTA-Net & 0.80 $\pm$ 0.52 & 32.66 $\pm$ 2.72 & 90.78 $\pm$ 3.68 \\
      & DL-SPIRiT & 0.39 $\pm$ 0.18 & 35.51 $\pm$ 2.39 & 92.39 $\pm$ 3.46 \\
      & VE-SDE & 0.33 $\pm$ 0.14 & 36.16 $\pm$ 2.53 & 92.22 $\pm$ 3.13 \\
      & DDS & 0.31 $\pm$ 0.15 & 36.59 $\pm$ 2.71 & 93.44 $\pm$ 2.83 \\
      & VP-SDE & 0.50 $\pm$ 0.21 & 34.35 $\pm$ 2.63 & 89.65 $\pm$ 3.87 \\
      & SPIRiT-Diffusion & \textbf{0.28 $\pm$ 0.20} & \textbf{37.22 $\pm$ 3.14} & \textbf{94.12 $\pm$ 2.73} \\

  \hline \hline
    \end{tabular}
  }
\end{table}

\subsection{Limitations}
Since SPIRiT-Diffusion is a coil-by-coil reconstruction method, its graphics memory footprint is much larger than that of conventional diffusion-based reconstruction methods. The memory needed depends on the number of receiver coils. In this study, the training data were compressed to 18 coils with an image size of 320 $\times$ 320, which required roughly 77GB of memory, while VE-SDE, based on CSM merging, required only 6GB of memory. Another issue is that SPIRiT-Diffusion generates coil images rather than a single image at each sampling step, aggravating the slow sampling procedure of diffusion models. Recent publications have reported that the optimal reverse variance has analytic forms and can achieve a 20 to 40 speed up compared to the full timesteps\cite{bao2022analytic, lu2022dpm}. This strategy is applicable to a variety of diffusion models, and we will investigate it to accelerate the sampling procedure of SPIRiT-Diffusion in our future work.

\section{Conclusions}\label{conclusions}
In this paper, we have proposed a novel paradigm of model-driven diffusion, where the diffusion equation is driven in accordance with the physics inherent in the optimization model. Specifically, we employed the SPIRiT model to drive a new diffusion model, enabling the accurate interpolation of missing $k$-space data. In comparison to conventional diffusion models formulated in the image domain, our approach demonstrated robust performance when confronted with challenges such as a limited FOV, phase singular points, and other factors that led to inaccurate CSM estimation. Finally, experimental validation on a 3D joint intracranial and carotid vessel wall imaging dataset confirmed the superiority of the proposed method.


\section*{Acknowledgement}
This study was supported in part by the National Key R\&D Program of China nos. 2020YFA0712200, 2021YFF0501402, National Natural Science Foundation of China under grant nos. 62322119, 12226008, 62125111, 62106252, 12026603, 62206273, 62106252, the Guangdong Basic and Applied Basic Research Foundation no. 2021A1515110540, Shenzhen Science and Technology Program under grant no. RCYX20210609104444089, JCYJ20220818101205012.

This study received support from Dr. Yue Xu and Dr. Hong Zhang of Sanya Central Hospital (Hainan Third People's Hospital), whose invaluable assistance with the prospective experimental scoring was instrumental to this research.

\appendix
\subsection{Estimating Score Functions}\label{Appendix:A}

According to the score-matching theory, under the expectation of $p(\mathbf{x}(t))$, we seek the score function $\mathbf{s}_{\boldsymbol{\theta}}$ to satisfy: 
\begin{equation}\begin{aligned}
    \mathbf{s}_{\boldsymbol{\theta}}(\mathbf{x}(t), t)=&\nabla_{\mathbf{x}(t)} \log p_{0 t}(\mathbf{x}(t) \mid \mathbf{x}(0))\\
    =&-\frac{\mathbf{x}(t)-\mathbf{x}(0)}{\boldsymbol{\Sigma}}
\end{aligned}\end{equation} 
Since $\boldsymbol{\Sigma^{\frac{1}{2}}}=\sigma \mathbf{S}\mathbf{S}^*$, by multiplying both sides of the equation by $\boldsymbol{\Sigma^{\frac{1}{2}}}$, we have 
\begin{equation}
    \sigma\mathbf{S}\mathbf{S}^*\mathbf{s}_{\boldsymbol{\theta}}(\mathbf{x}(t), t)=-\mathbf{z}
\end{equation}
where $\mathbf{z}$ is a vector following a standard normal distribution.
Continuing by multiplying both sides of the equation by $\mathbf{S}^*$, we get
\begin{equation}\begin{aligned}
    \sigma\mathbf{S}^*\mathbf{S}\mathbf{S}^*\mathbf{s}_{\boldsymbol{\theta}}(\mathbf{x}(t), t)=&-\mathbf{S}^*\mathbf{z}\\
    \sigma\mathbf{S}^*\mathbf{s}_{\boldsymbol{\theta}}(\mathbf{x}(t), t)=&-\mathbf{S}^*\mathbf{z}
\end{aligned}\end{equation} 
 where the second equation is a result of $\mathbf{S}^*\mathbf{S}=\mathbf{I}$.
Based on the above inference, the loss function for the score function $\mathbf{s}_{\boldsymbol{\theta}}$ can be defined as:
\begin{multline}
\begin{aligned}
        \boldsymbol{\theta}^{*}&=\underset{\boldsymbol{\theta}}{\arg \min } \mathbb{E}_{t}\big\{\lambda(t) \mathbb{E}_{\mathbf{x}(0)} \mathbb{E}_{\mathbf{x}(t) \mid \mathbf{x}(0)}\big[\big\|\sigma\mathbf{S}^*\mathbf{s}_{\boldsymbol{\theta}}(\mathbf{x}(t), t)\\&~~~~~~~~~~~~~~~~~~~~~~~~~~~~~~~~~~~~~~~~~~+\mathbf{S}^*\mathbf{z}\big\|_{2}^{2}\big]\big\}
\end{aligned}
\end{multline}

\subsection{The Forward SDE of SPIRiT-Diffusion}\label{Appendix:B}

Suppose \eqnref{SPIRiT iterating} is roughly interpreted as the discrete form of the reverse diffusion process conditioned on data consistency $\mathbf{Ax}-\mathbf{y}=\mathbf{n}$. Eliminating the impact of data consistency, and regarding the gradient of the self-consistency term as the drift coefficient $\mathbf{f}(\mathbf{x}, t)$ of the diffusion process, we express the corresponding forward discrete diffusion process as:
\begin{equation}\begin{aligned}
\mathbf{x}_{k+1}=\mathbf{x}_{k}+\eta_{k}\Psi(\mathbf{x}_k)+{\beta_{k}}\mathbf{z}_k 
\end{aligned}\end{equation} 
By introducing auxiliary variables $\bar{\eta}{k}=2T\beta{k}$ and $\bar{\beta}{k}=T\beta{k}^2$, the above equation can be rewritten as
\begin{equation}\begin{aligned}
\mathbf{x}_{k+1}=\mathbf{x}_{k}+\frac{\bar{\eta}_{k}}{2T}\Psi(\mathbf{x}_k)+\sqrt{\frac{\bar{\beta}_{k}}{T}}\mathbf{z}_k .
\end{aligned}\end{equation}
Without loss of generality, letting $\mathbf{x}(t)=\mathbf{x}_{k}$, $\mathbf{x}(t+\Delta t)=\mathbf{x}_{k+1}$, ${\beta}(t)=\bar{\beta}_{k}$, ${\eta}(t)=\bar{\eta}_{k}$ and $\Delta t=1/T$, we obtain:

\begin{equation}\begin{aligned}
\mathbf{x}(t+\Delta t)=\mathbf{x}(t)+\frac{\eta(t)\Delta t}{2}\Psi(\mathbf{x}(t))+\sqrt{\beta(t) \Delta t}\mathbf{z}(t).
\end{aligned}\end{equation}
Let $\Delta t=\frac{1}{T} \rightarrow 0$, \eqnref{init SPIRiT SDE} holds.

\bibliographystyle{IEEEtran}
\bibliography{refs}
\end{document}